\documentclass{svproc}
\usepackage{url}

\usepackage{graphicx}
\usepackage{float}
\usepackage{amsmath, amssymb}
\usepackage{booktabs, multirow, array}
\usepackage{xcolor}
\usepackage{tabularx}
\usepackage{colortbl}
\usepackage{placeins}
\usepackage{hyperref}
\usepackage{fontawesome5}
\usepackage{tikz}
\usepackage{marvosym}
\usetikzlibrary{shapes.geometric, arrows.meta, positioning, fit, backgrounds, calc}
\definecolor{staticblue}{RGB}{70, 130, 180}
\definecolor{agenticgreen}{RGB}{60, 140, 100}
\definecolor{artifactcolor}{RGB}{255, 223, 186}
\definecolor{contextcolor}{RGB}{198, 226, 255}
\definecolor{verifycolor}{RGB}{216, 242, 216}
\begin{document}
\mainmatter              
\title{From Features to Actions: Explainability in Traditional and Agentic AI Systems}
\titlerunning{Explainability in Traditional and Agentic AI Systems}
\author{
Sindhuja Chaduvula\inst{1} \and
Jessee Ho\inst{1} \and
Kina Kim\inst{2} \and
Aravind Narayanan\inst{1} \and
Ahmed Y. Radwan\inst{1} \and
Mahshid Alinoori\inst{1} \and
Muskan Garg\inst{3} \and
Dhanesh Ramachandram\inst{1} \and
Shaina Raza\inst{1}\textsuperscript{(\Letter)}
}
\authorrunning{S. Chaduvula et al.}
\tocauthor{S. Chaduvula, J. Ho, K. Kim, A. Narayanan, A. Y. Radwan,
           M. Alinoori, M. Garg, D. Ramachandram, S. Raza}
\institute{
Vector Institute for Artificial Intelligence, Toronto, Canada \\
\email{shaina.raza@vectorinstitute.ai}
\and
Independent Researcher
\and
Mayo Clinic, Rochester, MN, USA
}

\maketitle              

\begin{abstract}
Over the last decade, Explainable AI has primarily focused on interpreting individual model predictions, producing post-hoc explanations that relate inputs to outputs under a fixed decision structure.
Recent advances in large language models (LLMs) have enabled agentic AI systems whose behaviour unfolds over multi-step trajectories. In these settings, success and failure are determined by sequences of decisions rather than a single output. It remains unclear how explanation approaches designed for static predictions translate to agentic settings where behaviour emerges over time. In this work, we bridge this gap by comparing attribution-based explanations with trace-based diagnostics across both settings. 
Our results show that while attribution methods achieve stable feature rankings in static settings (Spearman $\rho = 0.86$), they cannot be applied reliably to diagnose execution-level failures in agentic trajectories. In contrast, trace-grounded rubric evaluation for agentic settings consistently localizes behaviour breakdowns and reveals that state tracking inconsistency is 2.7× more prevalent in failed runs and reduces success probability by 49\%. These findings motivate a shift towards trajectory-level explainability 
for evaluating and diagnosing autonomous AI behaviour in agentic systems. 
\keywords{Explainable AI, Agentic AI, Large Language Models, Trajectory-level Explainability} \\
\noindent\textbf{Resources:} \faGithub  \href{https://github.com/VectorInstitute/unified-xai-evaluation-framework}{ Code}
\quad
\faGlobe  \href{https://vectorinstitute.github.io/unified-xai-evaluation-framework/}{ Project Page}
\end{abstract}
    \section{Introduction}
    \label{sec:intro}
    
    Explainable Artificial Intelligence (XAI) has long struggled with a fundamental tension: as models grow more capable, their decision-making processes become increasingly opaque. The research community has responded with a rich toolkit of post-hoc explanation methods such as SHapley Additive exPlanations (SHAP)~\cite{lundberg2017unified}, Local Interpretable Model-agnostic Explanations (LIME)~\cite{ribeiro2016should}, saliency maps, and attention visualizations that infer model behaviour without modifying the underlying system~\cite{gunning2017xai,guidotti2018survey}, distinct from interpretability approaches that build transparency into models by design~\cite{linardatos2020explainable}. These techniques are well-suited to \textit{static prediction} settings, where a system's behaviour is evaluated with respect to a fixed input--output mapping.
    
    However, AI systems are increasingly becoming agentic~\cite{huang2025agentic}, particularly LLM-based agents that operate through sequences of observations, reasoning steps, and tool invocations rather than a single input--output prediction~\cite{farooq2025evaluating}. This shift challenges core assumptions of traditional explainability methods.

    \textbf{Problem Statement.} In static settings, explanations are commonly framed as attributions over input features or tokens for a single prediction. In agentic settings, many critical questions are trajectory-level: why an agent chose a tool, abandoned a strategy, propagated an incorrect state, or failed to recover after an error. In other words, the unit of explanation is no longer a single prediction but a \textit{trajectory}: a sequence of states, actions, and observations that collectively determines success or failure.
    
    We argue that bridging static and agentic explainability requires reframing explanation targets from feature-level influence to trajectory-level decision accounts, and coupling explanation artifacts with execution context and faithfulness signals grounded in the run. Explainability is also increasingly important in domains such as wireless communications and sensing, where AI systems operate in safety-critical environments. These applications further motivate the need for explainability frameworks that extend beyond static attribution toward temporally grounded reasoning and action tracing in adaptive agentic systems~\cite{yang2026movable,fan2026moeisac,illi2025ris}.

    \textbf{Contributions.} This paper makes three main contributions:
    \begin{itemize}
        \item We introduce a formal distinction between explainability for static predictors and explainability for agentic systems.
        \item We propose a cross-paradigm taxonomy of explanation targets and artifacts, from feature-level attributions to trajectory-level accounts.
        \item We empirically compare attribution-based explanations and trace-based diagnostics across (i) a static classification task and (ii) LLM-agent benchmarks (\texttt{TAU-bench Airline}~\cite{yao2024taubench}, \texttt{AssistantBench}~\cite{yoran2024assistantbench}), analyzing tool use, strategy shifts, and error recovery to motivate agent-centric requirements for actionable explanations.
    \end{itemize}
    
    \textbf{Scope.}
    We focus on explainability for \textit{agentic task completion} systems, where an LLM-based agent achieves a high-level goal through multi-step interaction (tool calls, intermediate reasoning, and state updates). Our emphasis is on \textit{post-hoc, trace-grounded} explanations and diagnostics derived from execution logs. We do not aim to provide mechanistic interpretability of model internals, nor to propose new agent training algorithms; instead, we study what explanation artifacts are informative and actionable under a unified static-vs-agentic framing.
    
    \textbf{Empirical Findings.} Our experiments on static classification tasks and agentic benchmarks (\texttt{TAU-bench Airline}~\cite{yao2024taubench}, \texttt{AssistantBench}~\cite{yoran2024assistantbench}) reveal a clear contrast between attribution-based and trace-based explanations. For static prediction tasks, SHAP and LIME produce stable feature rankings that capture aggregate correlates of model outputs. For agentic benchmarks, these attribution methods can still highlight which high-level behavioural dimensions (e.g., error recovery, tool correctness) correlate with task success in aggregate, but they do not reliably localize \textit{which constraint was violated} in a specific failed execution. In contrast, trace-based rubric evaluations grounded in execution logs consistently yield per-run, diagnostically actionable explanations by identifying violated behavioural constraints such as tool misuse, state inconsistency, or failed recovery. Overall, these findings empirically substantiate the paradigm gap highlighted by our taxonomy and motivate concrete requirements for explainability in agentic AI systems.
    
The rest of the paper is organized as follows:
Section~\ref{sec:related-work} reviews prior work. Section~\ref{sec:paradigms} formalizes the static versus agentic distinction and motivates trajectory-level explainability. Section~\ref{sec:exp-setup} describes the datasets, models, and evaluation setup. Section~\ref{sec:results} presents results comparing attribution-based methods and trace-grounded rubric analyses across both paradigms. Section \ref{sec:discussion} presents the discussion, limitations and future directions, and Section \ref{sec:conclusion} concludes the paper.

\section{Related Work}
\label{sec:related-work}
We situate our work within three areas: traditional explainability methods for static models, explainability techniques for LLMs and reasoning systems, and evaluation frameworks for agentic AI. 

\subsection{Traditional Explainability Methods}
\label{subsec:traditional-xai}
Traditional XAI methods have predominantly focused on \textit{extrinsic} (post-hoc) explanations, which seek to explain model behaviour after inference without modifying the underlying model, in contrast to \textit{intrinsic} interpretable models (e.g., linear models, decision trees)~\cite{doshi2017accountability}. Within the post-hoc paradigm, methods vary along two orthogonal dimensions: whether they are \textit{model-agnostic}, requiring only input--output access, or \textit{model-specific}, relying on gradients, activations, or internal representations. Broadly, post-hoc methods operate at the level of features, concepts, or internal mechanisms.

At the feature level, \textit{attribution} methods identify which input features or regions most strongly influence predictions. \textit{Model-agnostic} approaches include LIME~\cite{ribeiro2016should}, which fits local surrogate models to generate instance-level explanations, Partial Dependence Plots (PDPs) ~\cite{Friedman_2001}, which summarize marginal effects globally, and SHAP~\cite{lundberg2017unified}, which provides both model-agnostic (e.g., kernel-based) and model-specific variants. \textit{Model-specific} attribution methods, include \textit{gradient-based saliency}~\cite{simonyan2013deep},  \textit{Integrated Gradients} \cite{sundararajan2017_integratedgradients}, and \textit{Gradient-weighted Class Activation Mapping} (Grad-CAM)~\cite{selvaraju2016grad} for vision models, as well as Transformer analyses based on attention rollout~\cite{abnar2020quantifying} and relevance propagation~\cite{chefer2021transformer}. While widely used, feature-level attributions operate on high-dimensional input spaces (e.g. thousands of tokens or pixels) and often yield explanations that are difficult to interpret semantically.

\textit{Concept-based} methods address this limitation by shifting the unit of explanation from individual features to human-interpretable abstractions. Probing classifiers~\cite{alain2016understanding} assess the separability of concepts within hidden representations, while Testing with Concept Activation Vectors (TCAV)~\cite{kim2018interpretability} quantifies the influence of user-defined concepts on predictions. Recent work extends these ideas to vision-language models for concept-level analysis and intervention~\cite{parekh2024concept}. Although concept-based methods improve interpretability, they are typically correlational, identifying associations between representations and concepts without establishing how such concepts are causally implemented. 

\textit{Mechanistic interpretability} aims to move beyond correlational explainability by reverse-engineering internal computations and identifying causal circuits that produce specific behaviours and predictions~\cite{olah2020circuits}. This includes uncovering computational circuits in language models~\cite{wang2022interpretability}, automated causal discovery via Automated Circuit Discovery (ACDC)~\cite{conmy2023towards}, and extracting interpretable features using sparse autoencoders~\cite{cunningham2023sparse}. These approaches provide deeper insight into model internals but remain focused on explaining behaviour at a single inference step.

While useful, traditional XAI does not address how decisions unfold over time, how states change, or how agents adapt. This requires new explainability approaches for agentic systems.

\subsection{Explainability for LLMs and Reasoning}
\label{subsec:llm-xai}
The rise of LLMs has motivated post-hoc approaches that surface intermediate \textit{reasoning traces} (e.g., rationales and action logs) to improve transparency. Chain-of-thought (CoT)~\cite{wei2022cot} prompting elicits step-by-step rationales that often improve performance on complex tasks and can provide a human-readable account of the model's solution process. ReAct~\cite{yao2022react} interleaves reasoning traces with action execution, making tool use and decision steps more explicit. Reflexion~\cite{shinn2023reflexion} adds self-reflection and memory updates to support learning from mistakes over repeated trials. Most LLM ``explainability'' techniques in this space are post-hoc and often \textit{model-agnostic} at the interface level (e.g., prompting for rationales or reasoning traces), whereas attention visualization is \textit{model-specific} and requires access to internal signals.

In Retrieval-Augmented Generation (RAG) settings, explainability further requires tracing how retrieved evidence influences the final output and providing provenance over external sources. Attention visualization tools such as BertViz~\cite{vig2019bertviz} can reveal attention patterns over context, while recent work on knowledge-graph-based perturbations provides more structured provenance signals for RAG systems~\cite{balanos2025kgrag}. However, attention is not a faithful explanation in general, and provenance mechanisms are often necessary to support reliable attribution to sources.

A persistent concern is \textit{faithfulness}. CoT rationales are not guaranteed to reflect the causal factors driving model decisions and may be persuasive but misleading~\cite{jain2022attentionfaithfulness}. Counterfactual explanations~\cite{counterfactualexplanations2023} offer verification strategies, but these are rarely integrated into end-to-end agent workflows. This gap becomes evident in agentic settings, where explanations must account for sequences of actions and tool calls, not only a final text output.

\subsection{Evaluation Frameworks for Agentic AI}
\label{subsec:agent-eval-frameworks}

Several frameworks have recently emerged for evaluating agentic AI systems, particularly LLM-based agents that operate over multi-step trajectories. These approaches reflect growing recognition that agent behaviour must be assessed beyond final outputs and instead evaluated at the level of planning, tool use, and execution.
DeepEval~\cite{deepeval} decomposes agent behaviour into reasoning, action, and execution layers, using metrics such as plan quality, plan adherence, tool correctness, step efficiency, and task completion. While this decomposition enables target performance analysis, such predefined metrics may not transfer cleanly across tasks, domains, and agent designs. LangSmith \footnote{\url{https://www.langchain.com/}} supports final-response grading, trajectory comparison against reference workflows, and component-level testing, though its reliance on ``ideal'' reference traces can penalize alternative yet valid strategies. Ragas~\cite{ragas} introduces agent-oriented metrics for tool calling and goal fulfillment (e.g., Tool Call Accuracy/F1 and goal accuracy), supporting both reference-based and reference-free evaluation. Phoenix~\cite{arize} uses LLM judges to assess full tool-calling sequences, enabling end-to-end evaluation but offering limited diagnostic granularity for pinpointing where failures arise within a trajectory. Collectively, these frameworks reflect the growing recognition that agent evaluation must capture process-level behaviour. Yet they remain fragmented in metric definitions and focus primarily on \textit{performance} evaluation, rather than explicitly defining or evaluating \textit{explainability} as a first-class diagnostic artifact.

\textbf{Gap and Positioning.}
Prior work offers strong tools for explaining static predictors (e.g., attribution, concepts, circuits), and separate lines of work surface reasoning traces for LLMs or propose benchmarks and metrics for evaluating agent performance. Efforts to unify these through causal reasoning have been proposed~\cite{chakrabarty2025causal}, but remain largely conceptual and do not empirically evaluate explanation quality across paradigms. Our work bridges these areas by organizing XAI around a paradigm distinction (static vs.\ agentic) and empirically testing how explanations differ in their ability to diagnose failures in multi-step agents.

\section{Explainability Across Static and Agentic AI Paradigms}
\label{sec:paradigms}

XAI methods have evolved alongside advances in AI, from early rule-based systems and decision trees to post-hoc feature attributions, concept-based analyses, and mechanistic interpretability for deep neural networks. The rise of tool-using, multi-step \textit{agentic} systems introduces new explainability requirements: explanations must account for action sequences, tool interactions, and state evolution over time. In this section, we formalize the distinction between static and agentic paradigms, map existing XAI families through this lens, and introduce a unified framework, the Minimal Explanation Packet (MEP), for packaging explanation artifacts with context and verification when behaviour unfolds over trajectories. Figure~\ref{fig:mep-comparison} conceptually contrasts static and agentic MEPs, highlighting the shift from single input--output mapping to trajectory-level accounts. 

\begin{figure}[t]
\centering
\begin{tikzpicture}[
    scale=0.9,
    transform shape,
    node distance=0.35cm,
    box/.style={rectangle, draw, rounded corners=3pt, minimum width=4.2cm, 
                minimum height=1.2cm, align=center, font=\small},
    header/.style={rectangle, rounded corners=3pt, minimum width=4.6cm, 
                   minimum height=0.6cm, align=center, font=\small\bfseries, text=white},
    label/.style={font=\footnotesize\itshape, text=gray},
    arrow/.style={-{Stealth[length=2mm]}, thick, gray},
    dashed arrow/.style={-{Stealth[length=2mm]}, thick, gray, dashed}
]

\node[header, fill=staticblue] (static-header) {Static MEP};

\node[box, fill=artifactcolor, below=0.45cm of static-header] (static-artifact) 
    {SHAP, LIME and PDP feature scores\\[-1pt]{\scriptsize (token-level attribution)}};
\node[label, left=0.1cm of static-artifact, anchor=east] {Artifact};

\node[box, fill=contextcolor, below=0.35cm of static-artifact, minimum height=1.4cm] (static-context) 
    {Input text + predicted label\\[-1pt]{\scriptsize (single instance)}};
\node[label, left=0.1cm of static-context, anchor=east] {Context};

\node[box, fill=verifycolor, below=0.35cm of static-context, minimum height=1.2cm] (static-verify) 
    {Perturbation stability\\[-1pt]{\scriptsize (feature rank correlation)}};
\node[label, left=0.1cm of static-verify, anchor=east] {Verification};

\draw[arrow] (static-artifact) -- (static-context);
\draw[arrow] (static-context) -- (static-verify);

\node[font=\scriptsize, text=staticblue, below=0.25cm of static-verify] (static-ex) 
    {\textit{Example: Job category classifier}};

\node[header, fill=agenticgreen, right=2.2cm of static-header] (agentic-header) {Agentic MEP};

\node[box, fill=artifactcolor, below=0.45cm of agentic-header, minimum height=1.2cm] (agentic-artifact) 
    {Execution trace + reasoning\\[-1pt]{\scriptsize (tool calls, decisions)}};

\node[box, fill=contextcolor, below=0.35cm of agentic-artifact, minimum height=1.4cm] (agentic-context) 
    {Trajectory: $(s_0, a_0, o_0, \ldots, s_T)$\\[-1pt]
     {\scriptsize state snapshots, tool logs,}\\[-1pt]
     {\scriptsize retrieved docs, env feedback}};

\node[box, fill=verifycolor, below=0.35cm of agentic-context, minimum height=1.2cm] (agentic-verify) 
    {Rubric flags + replay checks\\[-1pt]
     {\scriptsize (intent, tool correctness,}\\[-1pt]
     {\scriptsize state consistency)}};

\draw[arrow] (agentic-artifact) -- (agentic-context);
\draw[arrow] (agentic-context) -- (agentic-verify);

\draw[dashed arrow, agenticgreen] 
    ($(agentic-context.east)+(0.1,0.2)$) 
    .. controls ($(agentic-context.east)+(0.6,0.5)$) 
    and ($(agentic-artifact.east)+(0.6,-0.3)$) .. 
    ($(agentic-artifact.east)+(0.1,-0.1)$);
\node[font=\tiny, text=agenticgreen, rotate=90] 
    at ($(agentic-context.east)+(0.7,0.6)$) {multi-step};

\node[font=\scriptsize, text=agenticgreen, below=0.25cm of agentic-verify] (agentic-ex) 
    {\textit{Example: Airline booking agent}};

\node[font=\footnotesize, text=gray] 
    at ($(static-artifact.east)!0.5!(agentic-artifact.west)$) {vs.};
\node[font=\footnotesize, text=gray] 
    at ($(static-context.east)!0.5!(agentic-context.west)$) {vs.};
\node[font=\footnotesize, text=gray] 
    at ($(static-verify.east)!0.5!(agentic-verify.west)$) {vs.};

\node[font=\small, align=center, 
      below=0.6cm of $(static-ex.south)!0.5!(agentic-ex.south)$] 
    {\textbf{Key shift:} Single input--output $\rightarrow$ Trajectory over time};

\end{tikzpicture}
\caption{MEP structure across static and agentic paradigms. Static MEPs pair 
a single input--output attribution with perturbation-based verification. Agentic 
MEPs extend this to trajectory-level context, replacing single predictions with 
tool calls, state snapshots, and rubric-based verification signals.}
\label{fig:mep-comparison}
\end{figure}

\begin{figure}[h]
    \centering
    \includegraphics[width=0.9\linewidth]{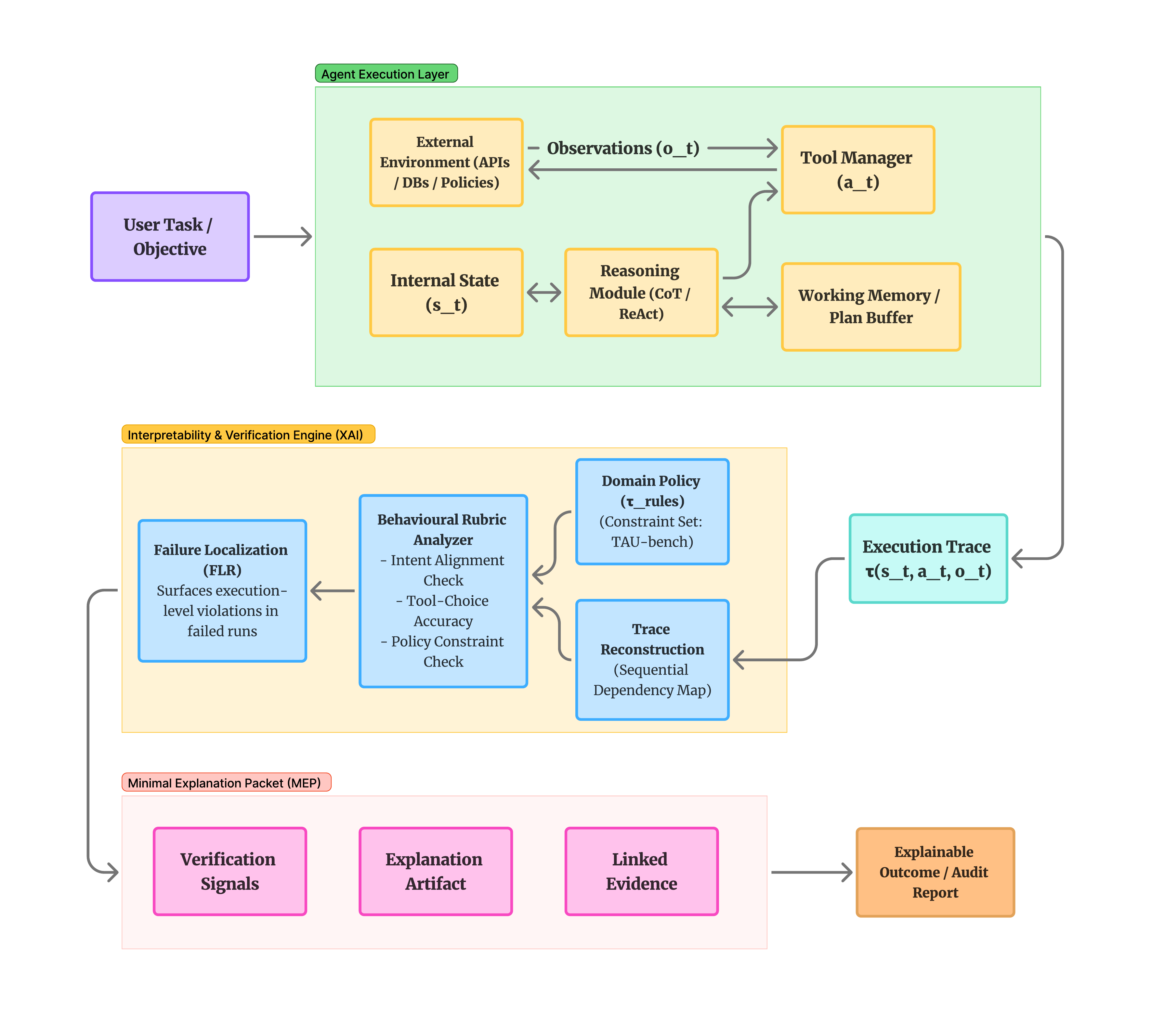}
    \caption{Agent execution loop with state--action--observation semantics. The 
    agent's reasoning module dispatches tool calls and updates internal state based 
    on environment observations. The XAI layer reconstructs execution traces and 
    applies behavioural rubrics to localize failures, producing a bundled MEP 
    output of artifact, linked evidence, and verification signals.}
    \label{fig:system-arch}
\end{figure}

\subsection{Paradigm Distinction}
\label{subsec:paradigm-distinction}

We distinguish between two paradigms that place fundamentally different demands on explanation. In the \textit{static (prediction-oriented) paradigm}, a system implements a fixed mapping $y = f(x)$ for a given input $x$, and explanations are defined with respect to a single input--output decision. This paradigm includes models with recurrence or iterative internal computation, provided that behaviour is evaluated only at the level of a single prediction. 

In \textit{agentic systems}, behaviour emerges as a trajectory $\tau = (s_0, a_0, o_0, s_1, a_1, \allowbreak o_1, \ldots, s_T)$, where the agent repeatedly observes the environment, reasons, and acts. Here $s_t$ denotes the internal state at step $t$, $a_t$ the action taken (including tool calls), and $o_t$ the observation received. Correctness and failure are defined at the level of the full trajectory rather than an individual decision.

This shift has two implications for explainability. First, scope expands from explaining single decisions to explaining full trajectories. Second, grounding shifts from input features to tool calls, state updates, retrieved evidence, and environmental feedback. Figure~\ref{fig:system-arch} illustrates the state--action--observation semantics underlying trajectory-level explainability, where tool interactions are coordinated within the reasoning module rather than by a separate reasoning entity.

\subsection{Mapping of XAI Methods Across Paradigms}
\label{subsec:taxonomy-table}

Table~\ref{tab:xai-unified} maps major families of explainability approaches across \textit{static prediction} and \textit{agentic} systems. We organize prior work by (i) the \textit{primary explanation target} (inputs/representations vs.\ trajectories/interactions) and (ii) the \textit{explanation artifact} produced (e.g., attributions, traces, counterfactuals, provenance, or verification signals). The table highlights how methods originally developed for single-step prediction must be adapted to support multi-step decision-making in agentic settings.


\paragraph{Attributions must become \textit{decision- and context-conditional}.}
In static prediction, attribution scores answer a well-posed question: which input features influenced a single output. In agentic systems, the analogous question is \textit{why this action now}, relative to available alternatives under the current state (memory, retrieved evidence, tool availability, and past observations). Step-level SHAP/LIME can be informative, but only when paired with (i) the agent’s \textit{action set} (tools/actions considered), (ii) the \textit{state variables} that condition the choice, and (iii) a mechanism for \textit{composing} local explanations into a trajectory-level account. Without this context, global summaries (e.g., PDP-style effects across runs) provide population-level correlates but systematically miss where and how failures arise within a specific trajectory.
\paragraph{Reasoning traces are necessary but insufficient.}
CoT- and ReAct-style traces expose intermediate reasoning and make multi-step structure legible to humans. However, they are \textit{self-reported} and can diverge from the actual causal drivers of behaviour, especially when tool results, memory updates, or retrieval events dominate subsequent decisions. Trace-based explanations therefore become more reliable when aligned with interaction logs, linking each stated intent to the corresponding action, tool output, and state update, and when discrepancies between ``said'' and ``done'' are surfaced explicitly.
\paragraph{Verification shifts from optional to first-class.}
In static XAI, faithfulness checks (e.g., perturbation stability) are often treated as optional add-ons. In agentic settings, failures can emerge from compounding effects across steps (state drift, cascading retrieval errors, or tool mis-specification), making verification essential. Practical signals include replay consistency under fixed seeds/tools, invariants over tracked state, and rubric-style flags that localize violated behavioural constraints. These checks convert explanations from plausible narratives into diagnostically actionable accounts.

\begin{table}[h]
\centering
\caption{Explainability approaches across static models and agentic systems.}
\label{tab:xai-unified}
\renewcommand{\arraystretch}{1.15}
\small
\setlength{\tabcolsep}{4pt}
\resizebox{\linewidth}{!}{
\begin{tabular}{|p{2.6cm}|p{3cm}|p{2.4cm}|p{3.6cm}|p{6.2cm}|}
\hline
\textbf{Family / target} &
\textbf{Representative methods} &
\textbf{Explanation artifact} &
\textbf{Role in static models} &
\textbf{Additional needs in agentic systems} \\
\hline

Attribution and saliency
& LIME, SHAP, PDP~\cite{ribeiro2016should,lundberg2017unified}; saliency, Grad-CAM~\cite{selvaraju2016grad}
& Feature scores, heatmaps
& Identify input regions/features that drive a single prediction
& Explain \textit{action selection} per step + connect attributions to tool choice + relate decisions to plan/state evolution \\
\hline

Attention-based analyses
& Attention rollout, relevance propagation~\cite{abnar2020quantifying,chefer2021transformer}
& Token influence paths
& Characterize token-level influence during generation
& Track attention shifts across steps, (planning $\rightarrow$ retrieval $\rightarrow$ execution), and across multiple context sources \\
\hline

Concept-based interpretability
& Probes, TCAV~\cite{hewitt2019structural,kim2018interpretability}; VLM concepts~\cite{parekh2024concept}
& Concept scores, probe accuracy
& Test whether concepts are encoded in representations
& Represent \textit{goals/subgoals}, constraints,  tool intent, and state variables, beyond raw feature concepts \\
\hline

Mechanistic interpretability
& Circuits, ACDC, sparse autoencoders~\cite{wang2022interpretability,conmy2023towards,cunningham2023sparse}
& Causal subgraphs, features
& Localize internal mechanisms behind predictions
& Analyze causal interactions across, memory modules, tool interfaces,  retrieval components, and policy updates \\
\hline

Reasoning traces and interaction logs
& CoT, ReAct, Reflexion~\cite{yao2022react,shinn2023reflexion}
& Stepwise rationales; tool logs
& Provide a human-readable rationale for one output
& Require trajectory-level linking: (reasoning $\rightarrow$ action $\rightarrow$ observation)  + replayable logs tied to outcomes \\
\hline

Evidence and provenance explanations
& Attention visualization, evidence paths~\cite{vig2019bertviz,balanos2025kgrag}
& Citations, provenance graphs
& Justify outputs using retrieved evidence
& Attribute \textit{which evidence} drove, which actions and revisions + detect retrieval-induced failure cascades\\
\hline

Counterfactual explanations
& Recourse, counterfactual evaluation~\cite{wachter2017counterfactual,counterfactualexplanations2023}
& What-if alternatives
& Identify minimal input changes to alter prediction
& Define counterfactuals over trajectories: alternative plans, tool calls,  and decision branches over time \\
\hline

Verification and faithfulness signals
& Simulatability, robustness checks~\cite{counterfactualsimulatability2023}
& Faithfulness signals
& Optional evaluation layer for explanation quality
& Make verification first-class: faithfulness checks over long horizons, consistency under replay, and rubric flags \\
\hline

Interactive / intervention-based explanations
& Causal probing via dialogue~\cite{kim2025because}
& Hypotheses with interventions
& Not applicable
& Explain multi-turn behaviour via, validated interventions on state, tool access, or observations \\
\hline

\end{tabular}}
\end{table}

\subsection{Evaluating Agentic Systems}
\label{subsubsec:agentic-metrics}

To compare explainability across paradigms, we use four criteria: (i) \textit{scope} (single-step vs.\ trajectory-level), (ii) \textit{grounding} (linkage to observable evidence such as inputs or traces), (iii) \textit{reliability/faithfulness} (signals that the explanation reflects the system’s actual behaviour), and (iv) \textit{auditability} (support for debugging, replay, and oversight). Table~\ref{tab:unified-criteria} summarizes how each criterion is operationalized in the two settings.

\begin{table}[t]
\centering
\caption{Unified criteria for evaluating explanations across static and agentic settings.}
\label{tab:unified-criteria}
\renewcommand{\arraystretch}{1.2}
\scriptsize
\begin{tabular}{|p{1.8cm}|p{2.6cm}|p{3.5cm}|p{3.7cm}|}
\hline
\textbf{Criterion} & \textbf{Definition} & \textbf{Static setting signal} & \textbf{Agentic setting signal} \\
\hline
Scope & Prediction-level vs.\ trajectory-level explanation & Local/global attribution for a single output & Trace-level explanation across actions, tools, and state updates \\
\hline
Grounding & Linkage to observable evidence & Input features, coefficients, saliency maps & Tool-call arguments, retrieved evidence, environment feedback, execution logs \\
\hline
Reliability/ faithfulness & Evidence that the explanation is trustworthy & Stability under perturbations; rank consistency & Replay-based checks; outcome-conditioned rubric statistics \\
\hline
Auditability & Support for debugging and oversight & Post-hoc feature inspection & Step-level replay, failure localization, tool correctness checks \\
\hline
\end{tabular}
\end{table}

We evaluate agentic explainability using trace-derived behavioural rubrics applied to complete execution trajectories. Our rubric evaluation builds on \textsc{Docent}~\cite{meng2025docent}, while our summary statistics follow the conditional analysis approach used in Holistic Agent Leaderboard 
evaluation harness (HAL-Harness)~\cite{hal2025}. For each run, an LLM-based judge assigns binary satisfaction/violation labels per rubric category using only the execution trace (actions, tool calls, observations, and intermediate state), without access to task outcomes or ground truth, thereby avoiding outcome leakage.

\paragraph{Notation.}
Let $i \in \{1,\dots,N\}$ index agent runs and $r \in \mathcal{R}$ index rubric categories (e.g., Intent Alignment, Tool Correctness). We define a binary violation indicator $f_{i,r} \in \{0,1\}$, where $f_{i,r}=1$ denotes a violation and $f_{i,r}=0$ indicates satisfaction. Each run has a binary task outcome $y_i \in \{0,1\}$, where $y_i=1$ denotes success and $y_i=0$ denotes failure.

\paragraph{Failure-mode prevalence.}
To quantify which violations are over-represented in failed runs, we compute:
\begin{equation}
P(f_r=1 \mid y=0), \qquad P(f_r=1 \mid y=1),
\end{equation}
and report their difference and ratio:
\begin{equation}
\Delta_{\text{prev}}(r) = P(f_r{=}1 \mid y{=}0) - P(f_r{=}1 \mid y{=}1),\qquad
\text{Ratio}_{\text{prev}}(r) = \frac{P(f_r{=}1 \mid y{=}0)}{P(f_r{=}1 \mid y{=}1)},
\end{equation}
where larger $\Delta_{\text{prev}}$ indicates violations more frequent in failures, and $\infty$ indicates violations exclusive to failed runs.

\paragraph{Reliability correlates.}
To assess how strongly a violation predicts reduced task success when it occurs, we compute:
\begin{equation}
P(y=1 \mid f_r=1), \qquad P(y=1 \mid f_r=0),
\end{equation}
and report:
\begin{equation}
\Delta_{\text{rel}}(r) = P(y{=}1 \mid f_r{=}1) - P(y{=}1 \mid f_r{=}0), \qquad
\text{RR}(r) = \frac{P(y{=}1 \mid f_r{=}1)}{P(y{=}1 \mid f_r{=}0)},
\end{equation}
where large negative $\Delta_{\text{rel}}$ indicates strong association with failure, and RR near $0$ indicates highly predictive failure signals.

\paragraph{Rubric categories.}
We use the rubric set in Table~\ref{tab:eval-metrics}, including Intent Alignment (goal-consistent actions), Plan Adherence (coherent multi-step planning), Tool Correctness (valid tool invocation), Tool-Choice Accuracy (appropriate tool selection), State Consistency (coherent state maintenance), and Error Recovery (detection and recovery from failures). All rubric prompts and scoring templates are released with our codebase.

\subsection{Minimal Explanation Packet (MEP)}
\label{subsec:mep-definition}

Agentic systems require explanations that are not standalone artifacts but bundles that connect behaviour to evidence and verification. We therefore introduce MEP, a lightweight, method-agnostic unit that packages:
\begin{enumerate}
    \item \textbf{Explanation artifact:} The human-interpretable explanation itself (e.g., feature attribution map, reasoning trace, tool-call summary).
    \item \textbf{Linked evidence and execution context:} Supporting material that grounds the artifact (e.g., input instance, execution trace, retrieved documents, tool-call logs, state snapshots).
    \item \textbf{Verification signals:} Indicators of explanation reliability (e.g., perturbation stability scores, rubric-based behavioural flags, replay-based consistency checks).
\end{enumerate}

\noindent The MEP is method-agnostic and can be instantiated at different scopes (local instance-level or global model-level), depending on the explanation goal.

\paragraph{Static MEP.}
For a classifier predicting job category from posting text:
\begin{itemize}
    \item \textit{Artifact:} SHAP feature attributions or LIME local explanations; PDPs when instantiated at a global (model-level) scope.
    \item \textit{Context:} Input text, predicted label, model confidence.
    \item \textit{Verification:} Rank correlation of top features across perturbed inputs.
\end{itemize}

\paragraph{Agentic MEP.}
For a tool-using agent completing an airline booking task:
\begin{itemize}
    \item \textit{Artifact:} Execution trace linking reasoning steps to actions (including tool calls and intermediate outputs).
    \item \textit{Context:} User request, observations at each step, tool arguments and returns, retrieved evidence, and state updates (e.g., memory/plan revisions).
    \item \textit{Verification:} Rubric-based behavioural flags (e.g., intent alignment, tool correctness, state consistency, error recovery) and replay-based consistency checks.
\end{itemize}

By shifting focus from isolated explanation artifacts to artifact-in-context with associated reliability evidence, the MEP supports auditing and replay rather than standalone narrative explanations. In our experiments (Section~\ref{sec:exp-setup}), we use MEP components to structure analysis of attribution- and trace-based explanations, emphasizing per-run diagnostic utility alongside aggregate correlations. For long agent trajectories, the Verification layer can evaluate temporally segmented execution windows to preserve trace fidelity and maintain localized causal attribution across extended interactions.
\section{Experimental Setup}
\label{sec:exp-setup}
We evaluate explanation quality under our unified framework in two complementary settings: (i) a \textit{static prediction} setting and (ii) an \textit{agentic} setting. Both are assessed using the same MEP-based criteria (Table~\ref{tab:unified-criteria}), such as \textit{scope}, \textit{grounding}, \textit{reliability/faithfulness}, and \textit{auditability}, while the concrete metrics reported in each setting are summarized in Table~\ref{tab:eval-metrics}.

\subsection{Datasets}
\label{subsec:datasets}

We use the \href{https://www.kaggle.com/datasets/madhab/jobposts}{Online Job Postings} dataset for binary text classification (IT vs.\ non-IT) to study traditional explainability in static prediction settings. 
For agentic behaviour, we evaluate two established tool-use benchmarks: (i) \textsc{TAU-bench Airline}~\cite{yao2024taubench}, which consists of structured airline customer-service tasks with API-mediated actions (e.g., search, rebook, cancel), and (ii) \textsc{AssistantBench}~\cite{yoran2024assistantbench}, which comprises web-based assistance tasks requiring multi-step navigation and information gathering.

\subsection{Models}
\label{subsec:models}
In the static setting, we use two lightweight classifiers: TF--IDF with Logistic Regression and a Text CNN baseline. In the agentic setting, we study tool-using LLM agents on \textsc{TAU-bench Airline} and \textsc{AssistantBench}, using o4-mini-2025-04-16~\cite{OpenAI2025o4mini} and GPT-4.1~\cite{openai2024gpt41} as the underlying models, respectively. All agent traces are labeled post-hoc with \textsc{Docent}~\cite{meng2025docent}, using a fixed GPT-5~\cite{openai2025gpt5systemcard} judge with medium reasoning effort for consistent rubric evaluation.

\textbf{Trace Collection}
Agent execution traces are collected using the HAL-Harness~\cite{hal2025}, which standardizes agent execution and logging across benchmarks, recording detailed trajectories for each run including sequences of actions, tool calls, and intermediate observations. These traces serve as the primary artifact for trajectory-level explainability analysis.

\subsection{Evaluation Metrics}
\label{subsec:eval-metrics}
Metrics for both static and agentic settings are summarized in Table~\ref{tab:eval-metrics}. 
In the static setting, we focus on the stability and consistency of attribution-based explanations under perturbations. 
In the agentic setting, we report rubric-based behavioural signals (discussed in Section \ref{subsubsec:agentic-metrics}) and analyze their relationship to task success and failure.  All rubric definitions, prompts, and scoring templates are released with our codebase. To reduce evaluator bias, the GPT-5 judge evaluates only reconstructed execution traces and rubric definitions without access to the agent’s internal states. Verification is therefore grounded in observable behavioral evidence, while the MEP framework also supports future integration of human and rule-based evaluation methods.

\subsection{Settings and Hyperparameters}
\label{subsec:hardware}
All experiments were conducted on CPU-only Linux servers. Agent traces for \textsc{AssistantBench} and \textsc{TAU-bench Airline} were generated using HAL-Harness and analyzed offline using \textsc{Docent}. Static explainability experiments were implemented in Python using standard scientific libraries. Table~\ref{tab:repro_hyperparams} reports the full set of software, model, and explanation settings used across both evaluations for reproducibility.

\section{Results and Analysis}
We first report static attribution results as a calibration baseline (Section~\ref{subsec:static-results}), then present trace-based rubric analyses for agentic benchmarks (Section~\ref{subsec:agentic-results}), and finally run a bridging experiment that projects trajectories into rubric features to compare attribution vs.\ trace-based explanations under a shared representation (Section~\ref{subsec:bridge}).
\label{sec:results}

\subsection{Static Explainability Results}
\label{subsec:static-results}
We evaluate traditional explainability methods in a static prediction setting using binary IT vs.\ non-IT classification on the Online Job Postings dataset. This experiment serves as a calibration baseline for explanation reliability in a regime where decisions are driven by sparse, semantically meaningful lexical features and explanations are naturally defined over a single input--output mapping.

\begin{table}[h]
\centering
\caption{Static setting: explanation stability (Spearman $\rho$) under perturbations. The results show that \textit{TF--IDF + Logistic Regression} achieves higher stability score compared to the Text CNN, indicating more consistent explanations under data perturbations.}
\label{tab:static_perf}
\small
\begin{tabular}{lc}
\toprule
\textbf{Model} & \textbf{Explanation Stability Score} \\
\midrule
\textbf{TF--IDF + Logistic Regression} & \textbf{0.8577} \\
Text CNN & 0.6127 \\
\bottomrule
\end{tabular}
\end{table}
As shown in Table~\ref{tab:static_perf}, TF--IDF + Logistic Regression yields substantially higher explanation stability (Spearman $\rho = 0.8577$) than the Text CNN ($\rho = 0.6127$), indicating more consistent attribution patterns under perturbations. We further illustrate representative local and global attribution behaviours in Figure~\ref{fig:lime-shap} (a)  for a single prediction using LIME, while panel (b) summarizes corpus-level feature effects using SHAP. Overall, while post-hoc methods provide useful insight into influential lexical features, they primarily explain \textit{final predictions} and offer limited visibility into intermediate reasoning steps or decision dynamics. 

\textbf{Key finding (static).}
Even under ideal conditions, static XAI methods explain outcomes at the prediction level and do not capture multi-step decision dynamics. This motivates trace-based evaluation in the agentic setting, where failures often arise from state updates, tool choices, and long-horizon trajectories rather than a single forward pass.

\subsection{Agentic Explainability}
\label{subsec:agentic-results}
We diagnose \textit{why} agents succeed or fail beyond aggregate accuracy 
using the execution traces and rubrics described in 
Section~\ref{subsec:models} and Section~\ref{subsubsec:agentic-metrics}.

\textbf{Overall Performance.}

Table~\ref{tab:agent_run_summary} shows that performance differs sharply across benchmarks: \textsc{TAU-bench Airline} achieves 56.0\% accuracy (28/50), while \textsc{AssistantBench} achieves 17.39\% (2/33). These benchmarks stress different skills, such as structured tool execution vs.\ open-ended web interaction, so accuracy alone can hide qualitatively different failure modes. This motivates trace-level rubric analysis. To minimize reconstruction artifacts, the trace reconstruction process preserves original execution logs, tool outputs, and observations without introducing inferred intermediate steps, ensuring that rubric evaluations remain grounded in observable trajectory evidence.

\textbf{Trace-Level Rubric Analysis.}
 
\begin{table}[t]
\centering
\caption{Agent evaluation runs for rubric-based analysis. Metrics are computed 
from HAL-Harness execution traces over all benchmark-defined task sets. 
\textbf{Bold} indicates best accuracy.}
\label{tab:agent_run_summary}

\scriptsize
\renewcommand{\arraystretch}{1.1}
\setlength{\tabcolsep}{3pt}

\begin{tabular}{p{2.3cm} p{2.2cm} p{1.5cm} c c c c c}
\toprule
Benchmark & Agent & Model & Tasks & Success & Failure & Acc. & Cost (\$) \\
\midrule

\textsc{AssistantBench}  
& Browser Agent 
& GPT-4.1 
& 33 
& 2 
& 31 
& 17.39\% 
& 14.15 \\

\textbf{TAU-bench Airline} 
& \textbf{TAU-bench Tool Calling} 
& \textbf{o4-mini-2025-04-16} 
& \textbf{50} 
& \textbf{28} 
& \textbf{22} 
& \textbf{56.00\%} 
& \textbf{11.36} \\

\bottomrule
\end{tabular}
\end{table}

\textit{Failure-mode prevalence.}
A rubric \textit{flag} indicates that a violation was detected for that run (i.e., the constraint was not satisfied).
Table~\ref{tab:failure_prevalence} reports how often each rubric violation appears in failed vs.\ successful runs. In \textsc{TAU-bench Airline}, failures are most strongly associated with \textit{State Tracking Consistency} ($\Delta = 0.333$, Ratio $= 2.7\times$), suggesting breakdowns that accumulate over long trajectories. In \textsc{AssistantBench}, \textit{Tool Choice Accuracy} stands out as a sparse but decisive failure mode that appears only in unsuccessful runs (Ratio $= \infty$).

\begin{table}[t]
\centering
\caption{Failure-mode prevalence per rubric, aggregated over all benchmark-defined 
tasks. $\Delta$~($\uparrow$) and Ratio~($\uparrow$) measure over-representation in 
failed runs. \textbf{Bold} and \underline{underline}: best and second-best rubrics 
(lowest $\Delta$); cell highlighting: worst failure mode (highest $\Delta$). 
$\dagger$ denotes holistic rubrics with the strongest failure association.}
\label{tab:failure_prevalence}

\scriptsize
\renewcommand{\arraystretch}{1.1}
\setlength{\tabcolsep}{3pt}

\begin{tabularx}{\textwidth}{p{2.4cm} X c c c c}
\toprule
Benchmark & Rubric & $P(\text{flag}\mid\text{failure})$ & $P(\text{flag}\mid\text{success})$ & $\Delta \uparrow$ & Ratio $\uparrow$ \\
\midrule

\textbf{TAU-bench Airline}
& Intent Alignment & 0.632 & 0.419 & 0.212 & 1.506 \\
& Error Awareness \& Recovery & 0.684 & 0.677 & $-0.007$ & 0.988 \\
& \cellcolor{red!20}State Tracking Consistency$^{\dagger}$
  & 0.526 & 0.194 & \cellcolor{red!20}0.333 & 2.719 \\
& \textbf{Tool Correctness} & \textbf{0.316} & \textbf{0.452} & \textbf{$-0.136$} & \textbf{0.699} \\
& \underline{Tool Choice Accuracy} & \underline{0.263} & \underline{0.290} & \underline{$-0.027$} & \underline{0.906} \\
& Plan Adherence & 0.211 & 0.129 & 0.081 & 1.632 \\

\midrule

\textbf{AssistantBench}
& \textbf{Intent Alignment} & \textbf{0.774} & \textbf{1.000} & \textbf{$-0.226$} & \textbf{0.77} \\
& Error Awareness \& Recovery & 0.516 & 0.500 & 0.016 & 1.03 \\
& \underline{State Tracking Consistency} & \underline{0.452} & \underline{0.500} & \underline{$-0.048$} & \underline{0.90} \\
& Tool Correctness & 0.548 & 0.500 & 0.048 & 1.10 \\
& \cellcolor{red!20}Tool Choice Accuracy
  & 0.484 & 0.000 & \cellcolor{red!20}0.484 & $\infty$ \\
& Plan Adherence & 0.032 & 0.000 & 0.032 & $\infty$ \\

\bottomrule
\end{tabularx}
\end{table}

\textit{Reliability correlates.}
Table~\ref{tab:reliability_correlates} measures predictive strength by comparing success rates when a violation is present vs.\ absent. In \textsc{TAU-bench Airline}, \textit{State Tracking Consistency} is the clearest predictor of failure (RR $= 0.51$), with success dropping by 36 percentage points when violated. This suggests that state drift accumulates silently until it causes irrecoverable errors. Interestingly, \textit{Tool Correctness} correlates positively with success (RR $= 1.24$), suggesting it captures minor, recoverable issues rather than fatal errors, as agents that trigger this flag may actually be attempting more complex tool interactions. In \textsc{AssistantBench}, most rubrics have limited predictive value due to the low-success regime, but \textit{Tool Choice Accuracy} and \textit{Plan Adherence} violations correspond to zero success (RR $= 0.00$), indicating these are hard blockers in web-based tasks. The MEP framework also enables localization of plan failures by linking rubric violations to specific reasoning steps, tool invocations, and intermediate execution states within the trajectory.

\begin{table}[h]
\centering
\caption{Reliability correlates per rubric, aggregated over all benchmark-defined 
tasks. $\Delta$ and RR~($\downarrow$) capture the absolute difference and relative 
change in success likelihood when a violation is present. \textbf{Bold} and 
\underline{underline}: highest and second-highest RR; cell highlighting: lowest RR. 
$\dagger$ denotes holistic rubrics with the strongest negative reliability effect.}
\label{tab:reliability_correlates}

\scriptsize
\renewcommand{\arraystretch}{1.1}
\setlength{\tabcolsep}{3pt}

\begin{tabularx}{\textwidth}{p{2.4cm} X c c c c}
\toprule
Benchmark & Rubric
& $P(\text{success}\mid\text{flag})$
& $P(\text{success}\mid\neg\text{flag})$
& $\Delta$
& RR $\downarrow$ \\
\midrule
\textbf{TAU-bench Airline}
& Intent Alignment & 0.52 & 0.72 & $-0.20$ & 0.72 \\
& Error Awareness \& Recovery & 0.618 & 0.625 & $-0.007$ & 0.99 \\
& \cellcolor{red!20}State Tracking Consistency$^{\dagger}$
  & 0.375 & 0.735 & $-0.36$ & \cellcolor{red!20}0.51 \\
& \textbf{Tool Correctness}
  & \textbf{0.70} & \textbf{0.567} & \textbf{0.133} & \textbf{1.24} \\
& \underline{Tool Choice Accuracy}
  & \underline{0.643} & \underline{0.611} & \underline{0.032} & \underline{1.05} \\
& Plan Adherence & 0.50 & 0.643 & $-0.143$ & 0.78 \\
\midrule
\textbf{AssistantBench}
& \textbf{Intent Alignment}
  & \textbf{0.077} & \textbf{0.000} & \textbf{0.077} & \textbf{$\infty$} \\
& Error Awareness \& Recovery & 0.059 & 0.062 & $-0.004$ & 0.94 \\
& \underline{State Tracking Consistency}
  & \underline{0.067} & \underline{0.056} & \underline{0.011} & \underline{1.20} \\
& Tool Correctness & 0.056 & 0.067 & $-0.011$ & 0.83 \\
& \cellcolor{red!20}Tool Choice Accuracy
  & 0.000 & 0.111 & $-0.111$ & \cellcolor{red!20}0.00 \\
& \cellcolor{red!20}Plan Adherence & 0.000 & 0.062 & $-0.062$ & \cellcolor{red!20}0.00 \\
\bottomrule
\end{tabularx}
\end{table}

\textbf{Key finding (agentic).}
Trace-level rubrics reveal \textit{how} failures unfold: \textsc{TAU-bench Airline} failures reflect gradual trajectory degradation (especially state inconsistency), whereas \textsc{AssistantBench} failures are driven by sparse but decisive mistakes. This explains why accuracy alone does not capture agent reliability.

\subsection{Bridging Static and Agentic Explainability}
\label{subsec:bridge}

Traditional XAI methods explain predictions from static input--output mappings, while agentic explainability targets behaviour unfolding over multi-step \textit{executions}. To compare these paradigms under a shared representation, we design a controlled bridging experiment on \textsc{TAU-bench Airline} execution traces.
Each trajectory is first labelled using \textsc{Docent} rubrics (Section~\ref{subsubsec:agentic-metrics}) and then encoded as a compact binary feature vector, where each dimension indicates whether a behavioural constraint is satisfied or violated. Using these rubric features, we train a logistic regression model to predict task success vs.\ failure (hyperparameters in Table~\ref{tab:repro_hyperparams}). We compute SHAP values using a linear explainer consistent with the model to quantify the influence of each rubric feature on the surrogate outcome predictor.

\begin{figure}[t]
    \centering
    \includegraphics[width=0.9\linewidth]{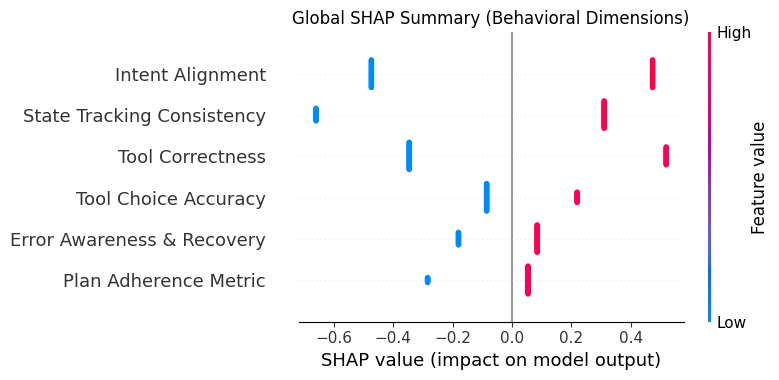}
    \caption{SHAP beeswarm plot for rubric-level outcome prediction on TAU-bench 
    Airline. Each point is one agent run; the x-axis shows contribution to predicted 
    success and color encodes feature value. Intent Alignment and State Tracking 
    Consistency show the widest spread, confirming their outsized influence on 
    outcomes. Tool Correctness skews positive, suggesting it flags recoverable 
    rather than fatal errors rather than fundamental reasoning breakdowns.}
    \label{fig:shap-beeswarm}
\end{figure}

\textbf{Attribution results.}
Table~\ref{tab:shap_rubric} reports mean absolute SHAP values, showing that \textit{Intent Alignment}, \textit{State Tracking Consistency}, and \textit{Tool Correctness} are the most influential predictors of outcome under this representation. Figure~\ref{fig:shap-beeswarm} visualizes the same trend: violations of these rubrics tend to push predictions toward failure, while satisfaction shifts predictions toward success.

\begin{table}[h]
\centering
\caption{Global SHAP attribution scores for the rubric-level outcome predictor (logistic regression). Higher values indicate greater overall influence. \textbf{Bold} and \underline{underline} denote the top two attributes.}
\label{tab:shap_rubric}
\small
\begin{tabular}{lc}
\toprule
\textbf{Rubric Attribute} & \textbf{Mean $|\mathrm{SHAP}|$} \\
\midrule
\textbf{Intent Alignment} & \textbf{0.473} \\
\underline{State Tracking Consistency} & \underline{0.422} \\
Tool Correctness & 0.415 \\
Tool Choice Accuracy & 0.122 \\
Error Awareness \& Recovery & 0.115 \\
Plan Adherence & 0.090 \\
\bottomrule
\end{tabular}
\end{table}

This experiment also shows that attribution methods can recover sensible \textit{global} importance rankings when agent executions are compressed into a low-dimensional, behaviourally grounded feature space. However, these attributions remain \textit{correlative}: they explain which rubric features drive the surrogate model’s outcome predictions, not what \textit{caused} a specific run to fail, consistent with recent evaluations of SHAP in reinforcement learning settings that highlight its limitations for causal interpretation~\cite{engelhart2024shaprl}. Table~\ref{tab:traditional_vs_agentic} summarizes this contrast, highlighting how attribution-based explanations over rubric features differ from trace-based agentic explainability even under a shared representation.

\textbf{Key finding (bridge).}
Projecting trajectories into rubric features makes SHAP useful for \textit{aggregate} analysis, but it does not provide trace-grounded diagnoses of \textit{where} or \textit{how} failures arise. Reliable explainability for long-horizon, tool-using agents therefore requires trajectory-level, trace-based reasoning, with attribution serving as a complementary summary tool.

\noindent\textbf{Overall takeaway.}
Static XAI is reliable when the task matches its assumptions, but it cannot explain multi-step failures. Trace-based rubrics diagnose trajectory failures directly, and rubric-to-SHAP bridging shows attribution can summarize global importance only after strong behavioural abstraction.

\section{Discussion}
\label{sec:discussion}
\subsection{Main Findings}
Overall, our results show a clear paradigm shift in what it means to \textit{explain} an AI system. In static prediction settings, feature attribution methods remain meaningful and reliable when the model's inductive bias aligns with task structure, as reflected by stable SHAP and LIME explanations in our classification experiments. However, when intelligence is expressed through multi-step interaction such as planning, tool use, state updates, and recovery, explanations tied to a single prediction no longer answer the questions practitioners actually ask: \textit{what failed, where, and why}~\cite{farooq2025evaluating}. Across \textsc{TAU-bench Airline} and \textsc{AssistantBench}, failures are governed by temporal and state-dependent phenomena, most prominently state inconsistency and compounding errors that attribution methods cannot localize, even when recovering sensible \textit{aggregate} correlations. In contrast, trace-grounded rubric analysis yields per-run, diagnostically actionable accounts that support auditing, debugging, and reliability assessment.

\textit{Error Analysis: Where Agentic Failures Arise}
Agentic failures rarely reduce to a single ``wrong output''. Instead, they arise from breakdowns in \textit{trajectory integrity}, where small deviations compound or a decisive branching mistake blocks progress. Our rubric analysis surfaces three recurring error classes.
In \textsc{TAU-bench Airline}, the strongest failure signal is \textit{State Tracking Consistency}. These failures manifest as latent divergence between the agent's evolving plan/memory and the environment state (e.g., stale constraints or mis-tracked entities), compounding across steps until a final tool call becomes irrecoverable -- a ``slow failure'' pattern where early steps appear reasonable but inconsistencies accumulate.
In \textsc{AssistantBench}, failures are dominated by sparse but decisive mistakes, most notably \textit{Tool Choice Accuracy} (and occasionally \textit{Plan Adherence}), corresponding to an incorrect interaction affordance after which recovery becomes unlikely under a fixed step budget -- a ``fast failure'' pattern where a single wrong branching decision can collapse the run.

Not all rubric flags are uniformly failure-predictive. \textit{Tool Correctness} can coincide with successful runs, suggesting some violations capture minor or recoverable issues rather than fundamental reasoning breakdowns; trace-based explanations still improve auditability by revealing \textit{what} went wrong even when it does not determine the outcome. Overall, error modes differ systematically by benchmark: \textsc{TAU-bench Airline} failures reflect \textit{gradual trajectory degradation} driven by state drift, whereas \textsc{AssistantBench} failures reflect \textit{hard blockers} driven by incorrect branching/tool choice, explaining why accuracy alone is insufficient and trace-grounded explanations are necessary to diagnose \textit{where} and \textit{how} failures occur.


\subsection{Limitations}
Our evaluation is limited in scope. We study a small set of tool-using LLM agents on \textsc{TAU-bench Airline} and \textsc{AssistantBench} using HAL-Harness. While these benchmarks capture realistic multi-step tool use, they do not cover the full range of agent architectures; our findings may not generalize to embodied agents, multi-agent coordination, or systems with online learning or persistent long-term memory.

Our trajectory-level explanations are derived post-hoc from execution traces and summarized using predefined behavioural rubrics. This abstraction enables scalable and consistent analysis but is necessarily coarse: it can hide fine-grained decision dynamics and supports primarily correlational rather than causal conclusions. Rubric labels are generated by an LLM judge via \textsc{Docent}, which introduces subjectivity despite fixed prompts and trace-only access, and relies on traces being complete, an assumption that may not hold under incomplete logging or partial observability.

Finally, \textsc{Docent} currently offers limited support for automated counterfactual interventions, causal validation, and direct integration with internal model representations, leaving these as directions for future work.


\section{Conclusion}
\label{sec:conclusion}
This work formalized the distinction between explainability in traditional predictive AI systems and the emerging requirements of multi-step agentic AI systems. We introduced the Minimal Explanation Packet (MEP), a structured framework designed to capture temporally grounded reasoning behavior, tool interactions, execution traces, and verification signals across complex agent trajectories.Through experiments on \textsc{TAU-bench Airline} and \textsc{ASSISTANTBENCH}, we demonstrated that conventional attribution-based explainability methods are insufficient for diagnosing failures in agentic workflows. In contrast, trace-based behavioral rubrics provide substantially improved visibility into planning failures, execution inconsistencies, and reasoning deviations that directly impact task success. Our findings suggest that explainability for agentic systems must evolve from static feature attribution toward trajectory-aware behavioral verification. While the proposed framework improves interpretability and debugging capabilities, several challenges remain open, including scalable human-grounded evaluation, robustness of automated judges, and explanation management for extremely long trajectories. Future work will explore hybrid human-AI verification strategies, adaptive trajectory summarization, and real-time explainability mechanisms for production-scale autonomous systems.

\subsubsection*{Acknowledgements}

Resources used in preparing this research were provided, in part, by the Province of Ontario and the Government of Canada through CIFAR, as well as companies sponsoring the Vector Institute (\url{http://www.vectorinstitute.ai/#partners}).

This research was funded by the European Union’s Horizon Europe research and innovation programme under the AIXPERT project (Grant Agreement No. 101214389), which aims to develop an agentic, multi-layered, GenAI-powered framework for creating explainable, accountable, and transparent AI systems.

\appendix
\section*{Appendix}

\begin{figure}[h]
    \centering
    \begin{minipage}{0.48\textwidth}
        \centering
        \includegraphics[width=\linewidth]{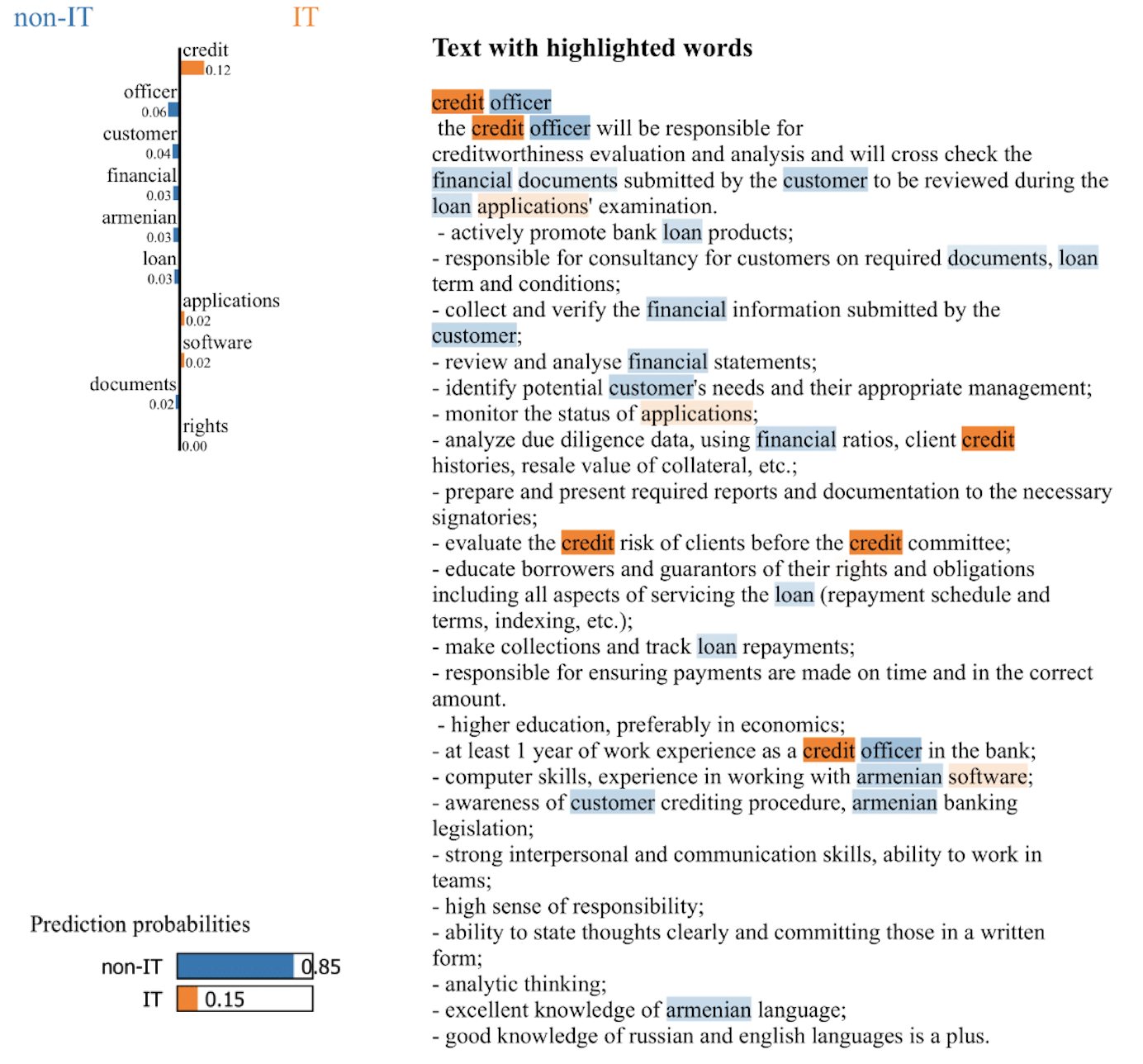}
        \label{fig:lime}
    \end{minipage}
    \hfill
    \begin{minipage}{0.48\textwidth}
        \centering
        \includegraphics[width=\linewidth]{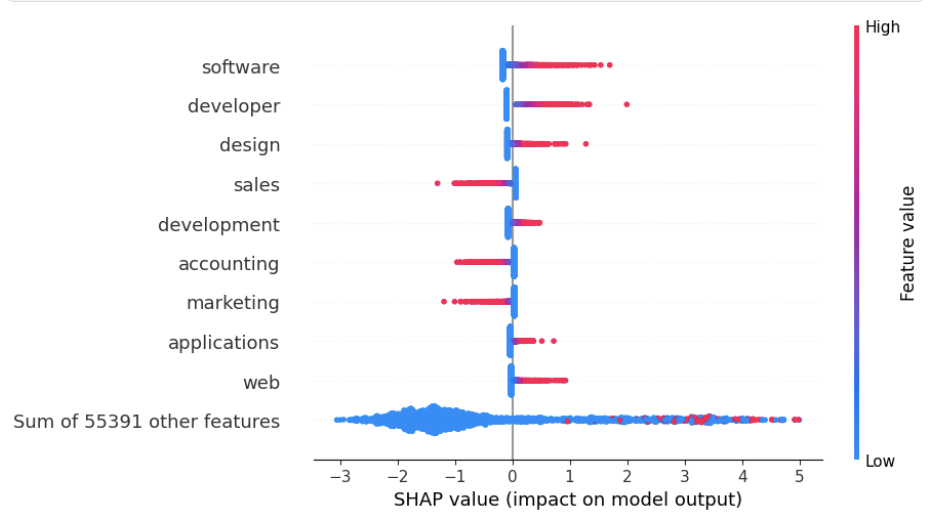}
        \label{fig:shap-global}
    \end{minipage}

    \begin{minipage}{0.48\textwidth}
        \scriptsize
        \textbf{(a) LIME explanation:}\\[2pt]
        Per-prediction feature importance for a single job posting. 
        Orange tokens push toward IT, blue toward non-IT, with 
        domain-specific terms carrying the strongest local signal.
    \end{minipage}
    \hfill
    \begin{minipage}{0.48\textwidth}
        \scriptsize
        \textbf{(b) SHAP global summary:}\\[2pt]
        Corpus-level feature effects across all predictions. 
        \textit{Software} consistently pushes toward IT and 
        \textit{accounting} toward non-IT, while ambiguous terms 
        like \textit{development} yield near-zero contributions.
    \end{minipage}
    \caption{Local (LIME) and global (SHAP) attribution for the static 
    IT classifier, illustrating instance-level and corpus-level feature 
    influence respectively.}
    \label{fig:lime-shap}
\end{figure}

\begin{table}[h]
\centering
\caption{Traditional attribution-based vs.\ trace-based agentic explainability under a shared rubric-derived representation.}
\label{tab:traditional_vs_agentic}
\renewcommand{\arraystretch}{1.15}
\scriptsize
\setlength{\tabcolsep}{3pt}
\begin{tabular}{p{3.6cm} p{3.9cm} p{3.9cm}}
\toprule
\textbf{Aspect} & \textbf{Traditional-XAI (SHAP/LIME)} & \textbf{Agentic-XAI (Docent)} \\
\midrule
Input representation
& Aggregated feature vector
& Full execution trace \\

Primary output
& Feature importance
& Rubric satisfaction/violation \\

Unit of explanation
& Outcome prediction
& Entire trajectory \\

Temporal reasoning
& $\times$
& $\checkmark$ \\

Tool/state awareness
& $\times$
& $\checkmark$ \\

Per-run failure localization
& Limited (indirect)
& Explicit (direct) \\

Explanation goal
& Correlative (\textit{what matters})
& Diagnostic (\textit{what went wrong}) \\
\bottomrule
\end{tabular}
\end{table}
\begin{table}[h]
\centering
\caption{Summary of experimental settings for both agentic and static explainability evaluations, including hardware, evaluation frameworks, benchmarks, models, training configurations, and explanation parameters used to ensure reproducibility.}
\label{tab:repro_hyperparams}
\scriptsize
\setlength{\tabcolsep}{4pt}
\begin{tabular}{p{3.4cm} p{8cm}}
\toprule
\textbf{Category} & \textbf{Setting / Value} \\
\midrule
\multicolumn{2}{l}{\textbf{Agentic Explainability Evaluation Setup}} \\
\midrule
Hardware & CPU-only Linux server (no GPU acceleration) \\
Python environment & Python 3.12; \texttt{scikit-learn}, \texttt{shap} \\
Agent evaluation harness & HAL-Harness~\cite{hal2025} (default benchmark configurations) \\
Trace analysis framework & \textsc{Docent}~\cite{meng2025docent} (post-hoc, trace-only evaluation) \\
LLM judge & GPT-5 (medium reasoning effort, single-pass, trace-only) \\
LLM hyperparameters & temperature $=0.1$; default sampling parameters \\ \\
Benchmarks & TAU-bench Airline~\cite{yao2024taubench} ($N=50$), AssistantBench~\cite{yoran2024assistantbench} ($N=33$) \\
Inference models & GPT-4.1 (AssistantBench), o4-mini-2025-04-16 (TAU-bench Airline) \\
Data usage & Full benchmark evaluation sets \\
\midrule

\multicolumn{2}{l}{\textbf{Static Explainability Evaluation Setup}} \\
\midrule
Compute Environment & Kaggle CPU-only Linux server (4 vCPUs, 16 GB RAM) \\
TF--IDF & N-gram range: (1,2); min\_df: 5; max\_df: 0.9; stopwords: English \\
Logistic Regression & max\_iter: 500; class\_weight: data-driven \\
Neural Model (CNN) & \begin{tabular}[t]{@{}l@{}}
Embedding: dim=100; mask\_zero=True; \\
Conv1D: filters=128, kernel\_size=5, activation=ReLU; \\
Pooling: GlobalMaxPooling1D; \\
Dense: 64 units, activation=ReLU; \\
Dropout: 0.5; \\
Output: 1 unit sigmoid
\end{tabular} \\
Data Split & 70\% train / 15\% validation / 15\% test \\
SHAP & Mean absolute SHAP value \\
LIME & Number of Features: 10 ; Random\_state: 42 \\
Saliency & Attribution method: Gradient × Input; Absolute score for ranking \\
\bottomrule
\end{tabular}
\end{table}
\begin{table}[h]
\centering
\caption{Evaluation metrics for static and agentic settings. Explanation stability is adopted from prior XAI robustness work. Agentic metrics are custom rubric signals defined in Section~\ref{subsubsec:agentic-metrics} and operationalized using Docent~\cite{meng2025docent}.}
\label{tab:eval-metrics}
\scriptsize
\renewcommand{\arraystretch}{1}
\setlength{\tabcolsep}{4pt}
\begin{tabular}{p{1.1cm} p{3cm}p{5.5cm} p{1.7cm} p{0.9cm}}
\toprule
\textbf{Setting} & \textbf{Metric} & \textbf{Description} & \textbf{MEP Criteria} & \textbf{Custom}\\
\midrule
Static & Explanation Stability &
Avg. Spearman rank correlation ($\rho$) across perturbed inputs or repeated runs  ~\cite{alvarez2018towards,hooker2019benchmark}&
Reliability &
$\checkmark$\\
\midrule
\multirow{6}{*}{Agentic}
& Intent Alignment &
Actions align with stated goals and task requirements &
Grounding &
$\checkmark$ \\
& Plan Adherence &
Maintains coherent multi-step plans throughout execution &
Grounding, Reliability &
$\checkmark$ \\
& Tool Correctness &
Invokes appropriate tools with valid parameters &
Auditability &
$\checkmark$ \\
& Tool-Choice Accuracy &
Selects optimal tools for given sub-tasks &
Grounding, Auditability &
$\checkmark$ \\
& State Consistency &
Maintains coherent internal state across steps &
Reliability &
$\checkmark$ \\
& Error Recovery &
Detects and recovers from execution failures &
Reliability &
$\checkmark$ \\
\bottomrule
\end{tabular}
\end{table}

\clearpage
\bibliographystyle{splncs03_unsrt}
\bibliography{fiaa-conference/references1}

\end{document}